\documentclass[11pt,a4paper]{article}
\usepackage{authblk}
\usepackage[hyperref]{emnlp2020}
\usepackage{siunitx}
\usepackage{latexsym}
\usepackage{times}
\usepackage{float}
\usepackage{footnote}
\usepackage{subcaption}
\usepackage{graphicx}
\usepackage{amsmath}
\usepackage{amssymb}
 \usepackage{booktabs}
 \usepackage{makecell}
 \usepackage{multirow}
\makesavenoteenv{tabular}
\makesavenoteenv{table}
\usepackage{graphicx}
\usepackage{hhline}
\usepackage{colortbl}
\usepackage{latexsym}
\usepackage{standalone}
\usepackage{url}
\usepackage{tikz}
\usepackage{soul}
\usepackage{fixltx2e}
\usepackage{color}
\usetikzlibrary{arrows,decorations.pathmorphing,backgrounds,positioning,fit,petri,calc}
\usepackage{pgfmath}

\definecolor{light-gray}{gray}{0.85}

\DeclareRobustCommand{\best}[1]{{\sethlcolor{gray}\hl{#1}}}
\DeclareRobustCommand{\second}[1]{{\sethlcolor{light-gray}\hl{#1}}}

\DeclareMathOperator{\MRR}{MRR}

\DeclareMathOperator{\R}{R}

\DeclareMathOperator*{\argmax}{arg\,max}
\DeclareMathOperator*{\argmin}{arg\,min}

\usepackage{microtype}

\aclfinalcopy 

\title{DagoBERT: Generating Derivational Morphology\\with a Pretrained Language Model}

\author[*$\ddag$]{Valentin Hofmann}
\author[$\dag$*]{Janet B. Pierrehumbert}
\author[$\ddag$]{Hinrich Sch\"utze}

\affil[*]{Faculty of Linguistics, University of Oxford}
\affil[$\dag$]{Department of Engineering Science, University of Oxford}
\affil[$\ddag$]{Center for Information and Language Processing, LMU Munich \protect\\ \texttt{valentin.hofmann@ling-phil.ox.ac.uk}}

\date{}

\begin{document}

\maketitle

\begin{abstract}
Can pretrained language models (PLMs) generate derivationally complex words?
We present the first study investigating this question, taking BERT as the example PLM.
We examine BERT's derivational capabilities in different settings, 
ranging from using the unmodified pretrained model to full finetuning.
Our best model, DagoBERT (Derivationally and generatively optimized BERT),
clearly outperforms the previous state of the art in derivation generation (DG).
Furthermore, our experiments show that the input segmentation crucially impacts 
BERT's derivational knowledge, suggesting that the performance of PLMs could be
  further improved if a morphologically informed vocabulary of
  units were used.
\end{abstract}

\section{Introduction}

What kind of linguistic knowledge is encoded by 
pretrained language models (PLMs) such as ELMo \citep{Peters.2018},
GPT-2 \citep{Radford.2019}, and BERT \citep{Devlin.2019}?
This question has attracted a lot of attention in NLP recently, 
with a focus on syntax (e.g., \citealp{Goldberg.2019})
and semantics (e.g., \citealp{Ethayarajh.2019}).
It is much less clear what PLMs learn about other aspects of language. 
Here, we present the first study on the knowledge of PLMs about derivational morphology, 
taking BERT as the example PLM.
Given an English cloze sentence such as \texttt{this jacket is
  \underline{\hspace{1cm}} .} and
a base such as \texttt{wear}, we ask: 
can BERT generate correct derivatives such as \texttt{unwearable}?

The motivation for this study is twofold. On the one hand, 
we add to the growing body of work on the linguistic capabilities of PLMs. 
Most PLMs segment words into subword units \citep{Bostrom.2020}, e.g., \texttt{unwearable} is segmented into \texttt{un}, \texttt{\#\#wear}, \texttt{\#\#able} by BERT's WordPiece tokenizer \citep{Wu.2016}. The fact that many of 
these subword units are derivational affixes suggests that PLMs might acquire 
knowledge about derivational morphology (Table \ref{tab:affixes-bert}), but this 
has not been tested. On the other hand, we are interested in derivation generation (DG) per se, a task that has been only addressed using LSTMs \citep{Cotterell.2017, Vylomova.2017, Deutsch.2018}, not models based on Transformers like BERT.

\begin{figure}[t!]
        \centering
        \begin{tikzpicture}
[xscale=0.9,
 thick,
 vec rectangle/.style={draw,rectangle,rounded corners=8pt,inner sep=2pt},
 vec circle/.style={draw,circle,minimum size=0.5cm,inner sep=0pt},
 model/.style={draw,rectangle,inner sep=6pt,rounded corners=6pt},
 token/.style={draw,rectangle,inner sep=0pt,minimum width=1.5cm,minimum height=0.75cm,font=\ttfamily}]

\pgfmathsetseed{1} 

\foreach \x / \xtext in {0/this,2/jacket,4/is,6/[MASK],8/wear,10/[MASK],12/.}
	\node [token] (tok-\x) at (\x + 0.375,1) {\xtext};

\foreach \x in {0,2,...,12}{
	\pgfmathsetmacro{\randfill}{random(20,100)}
	\node [vec circle,fill=white!\randfill!red] (c-0-\x-1) at (\x,2.5) {};
	\pgfmathsetmacro{\randfill}{random(20,100)}
	\node [vec circle,fill=white!\randfill!red] (c-0-\x-2) at (\x + 0.75,2.5){};
	\node [vec rectangle,fit=(c-0-\x-1) (c-0-\x-2)] (vec-0-\x) {};}
	
	
\foreach \x in {0,2,...,12}{
	\pgfmathsetmacro{\randfill}{random(20,100)}
	\node [vec circle,fill=white!\randfill!red] (c-2-\x-1) at (\x,5) {};
	\pgfmathsetmacro{\randfill}{random(20,100)}
	\node [vec circle,fill=white!\randfill!red] (c-2-\x-2) at (\x + 0.75,5){};
	\node [vec rectangle,fit=(c-2-\x-1) (c-2-\x-2)] (vec-2-\x) {};}

\node [vec circle,fill=white!20!red] (c-3-6-1) at (6-0.75,6.5) {};
\node [vec circle,fill=white!95!red] (c-3-6-2) at (6,6.5) {};
\node [vec circle,fill=white!85!red] (c-3-6-3) at (6+0.75,6.5){};
\node [vec circle,fill=white!98!red] (c-3-6-4) at (6 + 1.5,6.5){};
\node [vec rectangle,fit=(c-3-6-1) (c-3-6-2) (c-3-6-3) (c-3-6-4)] (vec-3-6) {};
	
\node [vec circle,fill=white!75!red] (c-3-10-1) at (10-0.75,6.5) {};
\node [vec circle,fill=white!97!red] (c-3-10-2) at (10,6.5) {};
\node [vec circle,fill=white!20!red] (c-3-10-3) at (10+0.75,6.5){};
\node [vec circle,fill=white!80!red] (c-3-10-4) at (10 + 1.5,6.5){};
\node [vec rectangle,fit=(c-3-10-1) (c-3-10-2) (c-3-10-3) (c-3-10-4)] (vec-3-10) {};

\foreach \x in {0,12}{
\node [vec circle,draw=none] (c-3-\x-1) at (\x,6.5) {};
\node [vec circle,draw=none] (c-3-\x-2) at (\x + 0.75,6.5){};
\node [vec rectangle,fit=(c-3-\x-1) (c-3-\x-2),draw=none] (vec-3-\x) {};}

\foreach \x / \xtext in {6/un,10/\#\#able}
	\node [token] (mask-\x) at (\x + 0.375,8) {\xtext};
	
\foreach \x in {0,2,...,12}
	\path (tok-\x) edge[->,dashed] (vec-0-\x);
	
\foreach \source in {0,2,...,12}
	\foreach \dest in {0,2,4,8,12}
		\path (vec-0-\source) edge[->,out=90,in=-90,gray!50!white] (vec-2-\dest);
	
\foreach \source in {0,2,...,12}
	\foreach \dest in {6,10}
		\path (vec-0-\source) edge[->,out=90,in=-90] (vec-2-\dest);
%
%
		
\foreach \x in {6,10}
	\path (vec-2-\x) edge[->] (vec-3-\x);
	
\foreach \x in {6,10}
	\path (vec-3-\x) edge[->,dashed] (mask-\x);
	
\begin{scope}[on background layer]	
\node [model,fit=(vec-0-0) (vec-2-12),blue!50,fill=blue!5,very thick,
label={[rotate=-90,xshift=-0.7cm,yshift=0.25cm,blue!50]right:\large BERT}] {};
\node [model,fit=(vec-3-0) (vec-3-12),orange!50,fill=orange!5,very thick,
label={[rotate=-90,xshift=-0.475cm,yshift=0.25cm,orange!50]right:\large DCL}] {};
\end{scope}

\end{tikzpicture}
        \caption{Basic experimental setup. We input
          sentences such as \texttt{this jacket is
            unwearable .} to BERT, mask out derivational
          affixes, and recover them using a derivational
          classification layer (DCL).}  \label{fig:setup}
\end{figure}
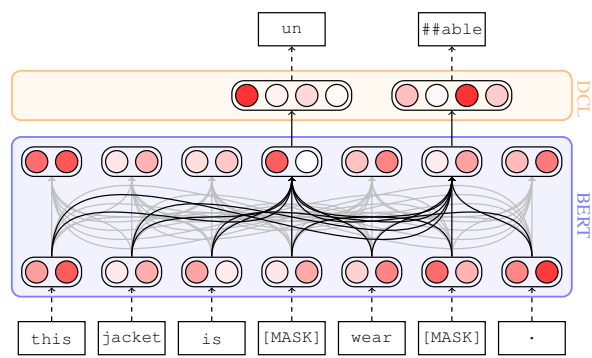

\begin{table} [t!]\centering
\resizebox{\linewidth}{!}{%
\begin{tabular}{@{}ll@{}}
\toprule
Type & Examples \\
\midrule
Prefixes  & \makecell[l]{\texttt{anti}, \texttt{auto}, \texttt{contra}, \texttt{extra}, \texttt{hyper}, \texttt{mega},\\ \texttt{mini}, \texttt{multi}, \texttt{non}, \texttt{proto}, \texttt{pseudo} }  \\
Suffixes  & \makecell[l]{ \texttt{\#\#able}, \texttt{\#\#an}, \texttt{\#\#ate}, \texttt{\#\#ee}, \texttt{\#\#ess}, \texttt{\#\#ful},\\ \texttt{\#\#ify}, \texttt{\#\#ize}, \texttt{\#\#ment}, \texttt{\#\#ness},  \texttt{\#\#ster}} \\

\bottomrule
\end{tabular}}
\caption{Examples of derivational affixes in the BERT WordPiece vocabulary. Word-internal WordPiece tokens are marked with \texttt{\#\#} throughout the paper.}  \label{tab:affixes-bert}
\end{table}

\textbf{Contributions.} We develop the first framework for generating derivationally complex English words with a PLM, specifically BERT, and analyze BERT's performance in different settings. Our best model, DagoBERT (Derivationally and generatively optimized BERT), clearly outperforms an LSTM-based model, the previous state of the art. We find that DagoBERT's errors are mainly due to syntactic and semantic overlap between affixes.
Furthermore, we show that the input segmentation impacts how much derivational knowledge is available to BERT, both during training and inference. This suggests that the performance of PLMs could be
  further improved if a morphologically informed vocabulary of
  units were used.
We also publish the largest dataset of derivatives in context to date.\footnote{We make our code and data publicly available at \url{https://github.com/valentinhofmann/dagobert}.}

\section{Derivational Morphology} \label{sec:deriv-morph}

Linguistics divides morphology into inflection and derivation. Given a lexeme such as \texttt{wear}, while inflection produces word forms such as \texttt{wears}, derivation produces new lexemes such as \texttt{unwearable}. There are several differences between inflection and derivation \citep{Haspelmath.2010}, two of which are particularly important for the task of DG.\footnote{It is important to note that the distinction between inflection and derivation is fuzzy \citep{Hacken.2014}.}

First, derivation covers a much larger spectrum of meanings
than inflection \citep{Acquaviva.2016}, and it is not
possible to predict in general with which of them a
particular lexeme is compatible. This is different from inflectional paradigms, where it is automatically
clear whether a certain form will exist \citep{Bauer.2019}. Second, the relationship between form and meaning is more varied in derivation than inflection. On the one hand, derivational affixes tend to be highly polysemous, i.e., individual affixes can represent a number of related meanings \citep{Lieber.2019}. On the other hand, several affixes can represent the same meaning, e.g., \texttt{ity} and \texttt{ness}. While such competing affixes are often not completely synonymous as in the case of \texttt{hyperactivity} and \texttt{hyperactiveness}, there are examples like \texttt{purity} and \texttt{pureness} or \texttt{exclusivity} and \texttt{exclusiveness} where a semantic distinction is more difficult to gauge \citep{Bauer.2013, Plag.2020}. These differences make 
learning functions from meaning to form harder for derivation than inflection.

Derivational affixes differ in how productive they are, i.e., how readily 
they can be used to create new lexemes \citep{Plag.1999b}. 
While the suffix \texttt{ness}, e.g., can attach to practically all English 
adjectives, the suffix \texttt{th} is much more limited in its scope
of applicability. In this paper, we focus on productive affixes such as \texttt{ness}
and exclude unproductive affixes such as \texttt{th}.
Morphological productivity has been the subject of much work in psycholinguistics 
since it reveals implicit cognitive generalizations (see \citet{Dal.2016} 
for a review), making it an interesting phenomenon
to explore in PLMs. Furthermore, 
in the context of NLP applications such as sentiment analysis,
productively formed derivatives 
are challenging because they tend to have very low frequencies and
often only occur once (i.e., they are hapaxes) or 
a few times in large corpora \citep{Mahler.2017}. 
Our focus on productive derivational morphology has crucial consequences 
for dataset design (Section \ref{sec:data}) and model evaluation 
(Section \ref{sec:experiments}) in the context of DG.

\section{Dataset of Derivatives} \label{sec:data}

\begin{table*} [t!]\centering
\resizebox{\linewidth}{!}{%
\begin{tabular}{@{}lrrrlrrlrrl@{}}
\toprule
{} & {} & \multicolumn{3}{c}{P}  &  \multicolumn{3}{c}{ S } &  \multicolumn{3}{c}{ PS } \\
\cmidrule(lr){3-5}
\cmidrule(lr){6-8}
\cmidrule(l){9-11}
Bin & $\mu_f$ & $n_d$ & $n_s$ & Examples & $n_d$ & $n_s$ & Examples & $n_d$ & $n_s$ &  Examples \\
\midrule
B1 & .041 & 60,236 & 60,236 & \texttt{antijonny}  & 39,543 & 39,543 & \texttt{takeoverness} & 20,804 & 20,804 & \texttt{unaggregateable} \\
B2 & .094 & 39,181 & 90,857 & \texttt{antiastronaut} & 22,633 & 52,060 & \texttt{alaskaness} & 8,661 & 19,903 & \texttt{unnicknameable} \\
B3 & .203 & 26,967 & 135,509 & \texttt{antiyale} & 14,463 & 71,814 & \texttt{blockbusterness} & 4,735 & 23,560 & \texttt{unbroadcastable} \\
B4 & .423 & 18,697 & 196,295 & \texttt{antihomework} & 9,753 & 100,729 & \texttt{abnormalness} & 2,890 & 29,989 &  \texttt{unbrewable} \\
B5 & .868 & 13,401 & 287,788 & \texttt{antiboxing} & 6,830 & 145,005 & \texttt{legalness} & 1,848 & 39,501 & \texttt{ungooglable}\\
B6 &1.750& 9,471 & 410,410 & \texttt{antiborder} & 4,934 & 211,233 & \texttt{tragicness} & 1,172 & 50,393 & \texttt{uncopyrightable} \\
B7 &3.515& 6,611 & 573,442 & \texttt{antimafia} & 3,580 & 310,109 & \texttt{lightweightness} & 802 & 69,004 & \texttt{unwashable}  \\
\bottomrule
\end{tabular}}
\caption{Data summary statistics. The table shows statistics of the data used in the study by frequency bin and affix type. We also provide example derivatives with \texttt{anti} (P), \texttt{ness} (S), and \texttt{un\#\#able} (PS) for the different bins. $\mu_f$: mean frequency per billion words; $n_d$: number of distinct derivatives; $n_s$: number of context sentences.}  \label{tab:bins}
\end{table*}

We base our study on a new dataset of derivatives in context similar in form to the one released by \citet{Vylomova.2017},
i.e., it is based on sentences with a derivative (e.g., \texttt{this jacket is unwearable .}) that are altered by masking 
the derivative (\texttt{this jacket is
  \underline{\hspace{1cm}} .}).
Each item in the dataset consists of (i) the altered
sentence, (ii) the derivative (\texttt{unwearable}) and (iii)
the base (\texttt{wear}).
The task is to generate the correct
derivative given the altered sentence and the base.
We use sentential contexts rather than tags to represent derivational meanings because they better reflect the semantic variability inherent in derivational morphology (Section \ref{sec:deriv-morph}). While \citet{Vylomova.2017} use Wikipedia,
we extract the dataset from Reddit.\footnote{We 
draw upon the entire Baumgartner Reddit
Corpus, a collection of all public Reddit posts 
available at \url{https://files.pushshift.io/reddit/comments/}.} Since productively formed derivatives are not part of the language norm initially \citep{Bauer.2001}, social media is a particularly fertile ground for our study.

For determining derivatives, we use the algorithm introduced by \citet{Hofmann.2020a}, 
which takes as input a set of prefixes, suffixes, and bases and checks for each word in the data whether it can be derived from a base using a combination of prefixes and suffixes. The algorithm is
sensitive to morpho-orthographic rules of English
\cite{Plag.2003}, e.g., when \texttt{ity} is removed from
\texttt{applicability}, the result is \texttt{applicable}, not
\texttt{applicabil}. Here, we use BERT's prefixes, suffixes, and bases as input to the algorithm. 
Drawing upon a comprehensive list of 52 productive prefixes and 49 productive suffixes in English \citep{Crystal.1997}, we find that 48 and 44 of them are contained in BERT's vocabulary. 
We assign all fully alphabetic words with more than 3 characters in BERT's vocabulary except for stopwords and previously identified affixes to the set of bases, yielding a total of 20,259 bases. 
We then extract every sentence including a word that is derivable
from one of the bases using at least one of the prefixes
or suffixes from all publicly available Reddit posts.

The sentences are filtered to contain between 10 and 100 words, i.e., they provide more contextual information than the example sentence above.\footnote{We also extract the preceding and following sentence for future studies on long-range dependencies in derivation. However, we do not exploit them in this work.} See Appendix \ref{app:preproc} for details about data preprocessing. The
resulting dataset comprises 413,271 distinct derivatives in
123,809,485 context sentences, making it more than two
orders of magnitude larger than the one released by
\citet{Vylomova.2017}.\footnote{Due to the large number of
  prefixes, suffixes, and bases, the dataset can be valuable
  for any study on derivational morphology, irrespective of
  whether or not it focuses on DG.} To get a sense of segmentation errors in the dataset, we randomly pick 100 derivatives for each affix and manually count missegmentations. We find that the average precision of segmentations in the sample is .960$\pm$.074, with higher values for prefixes (.990$\pm$.027) than suffixes (.930$\pm$.093).

For this study, we extract all derivatives with a frequency $f \in [1, 128)$
  from the dataset. We divide the derivatives into 7
  frequency bins with $f = 1$ (B1),  $f \in [2, 4)$ (B2),
    $f \in [4, 8)$ (B3), $f \in [8, 16)$ (B4),  $f \in [16,
          32)$ (B5), $f \in [32, 64)$ (B6), and $f \in [64,
              128)$ (B7). Notice that we focus on low-frequency derivatives since we are interested in 
productive derivational morphology (Section \ref{sec:deriv-morph}). 
In addition, BERT is likely to have seen high-frequency derivatives multiple times during pretraining and might be able to predict the affix because it has memorized the connection between the base and the affix, not because it has knowledge of derivational morphology. BERT's pretraining corpus has 3.3 billion words, i.e., words in the lower frequency bins are very unlikely to 
have been seen by BERT before. This observation also holds for average speakers of English, who have been shown to encounter at most a few billion word tokens in their lifetime \citep{Brysbaert.2016b}.

Regarding the number of affixes, we confine ourselves to three cases:
derivatives with one prefix (P), derivatives with one suffix
(S), and derivatives with one prefix and one suffix (PS).\footnote{We denote affix bundles, i.e., combinations of prefix and suffix, by juxtaposition, e.g., \texttt{un\#\#able}.} We treat these cases separately because they are known to have 
different linguistic properties. 
In particular, since suffixes in English can change the POS of a lexeme, the syntactic context is more affected by suffixation than by prefixation.
Table \ref{tab:bins} provides summary statistics for the seven frequency bins as well as example derivatives for P, S, and PS. For each bin, we randomly split
              the data into 60\% training, 20\% development,
              and 20\% test. Following
              \citet{Vylomova.2017}, we distinguish the
              lexicon settings
SPLIT (no overlap between bases in train, dev, and test) and
SHARED (no constraint on overlap).

\section{Experiments} \label{sec:experiments}

\subsection{Setup} \label{sec:setup}

To examine whether BERT can generate derivationally complex
words, we use a cloze test: given a sentence with a masked
word such as \texttt{this jacket is \underline{\hspace{1cm}}
  .} and a base such as \texttt{wear}, the task is to
generate the correct derivative such as
\texttt{unwearable}. The cloze setup has been previously used in psycholinguistics to probe derivational morphology \citep{Pierrehumbert.2006, Apel.2011} and was introduced to
NLP in this context by \citet{Vylomova.2017}.

In this work, we frame DG as an affix classification task, i.e., we predict which affix is most likely to occur with a given base in a given
context sentence.\footnote{In the case of PS, we predict which affix bundle (e.g., \texttt{un\#\#able}) is most likely to occur.} More formally, given a base $b$ and a context sentence $\mathbf{x}$ split into left and right contexts $\mathbf{x}^{(l)} = (x_1, \dots, x_{d-1})$ and $\mathbf{x}^{(r)} = (x_{d+1}, \dots, x_n)$, with $x_d$ being the masked derivative, we want to find the affix $\hat{a}$ such that 
\begin{equation}
\hat{a} = \argmax_a P \left( \psi(b, a) | \mathbf{x}^{(l)}, \mathbf{x}^{(r)} \right),
\end{equation} 
where $\psi$ is a function mapping bases and affixes onto
derivatives, e.g., $\psi(\text{\texttt{wear}},
\text{\texttt{un\#\#able}}) = \text{\texttt{unwearable}}$.
Notice we do not model the function $\psi$ itself, i.e., we only predict derivational categories, not the morpho-orthographic changes that accompany their realization in writing. One reason for this is that as opposed to previous work, our study focuses on low-frequency derivatives, for many of which $\psi$ is not right-unique, e.g., \texttt{ungoogleable} and \texttt{ungooglable} or \texttt{celebrityness} and \texttt{celebritiness} occur as competing forms in the data.

As a result of the semantically diverse nature of derivation
(Section \ref{sec:deriv-morph}), deciding whether a
particular prediction $\hat{a}$ is correct or not is less
straightforward than it may seem. Taking again the example
sentence \texttt{this jacket is \underline{\hspace{1cm}} .}
with the masked derivative \texttt{unwearable}, compare the
following five predictions:
\setlength{\leftmargini}{1.5em}
\begin{itemize}
\item[--] $\psi(b, \hat{a}) = \text{\texttt{wearity}}$: ill-formed;
\item[--] $\psi(b, \hat{a}) = \text{\texttt{wearer}}$: well-formed, syntactically incorrect (wrong POS);
\item[--] $\psi(b, \hat{a}) = \text{\texttt{intrawearable}}$: well-formed, syntactically correct, semantically dubious;
\item[--] $\psi(b, \hat{a}) = \text{\texttt{superwearable}}$: well-formed, syntactically correct, semantically possible, but did not occur in the example sentence;
\item[--] $\psi(b, \hat{a}) = \text{\texttt{unwearable}}$: well-formed, syntactically correct, semantically possible, and did occur in the example sentence.
\end{itemize}
These predictions reflect increasing degrees of derivational knowledge. A priori, where to draw the line between correct and incorrect predictions on this continuum is not clear, especially with respect to the last two cases. Here, we apply the most conservative criterion: a prediction $\hat{a}$ is only judged correct if $\psi(b, \hat{a}) = x_d$, i.e., if $\hat{a}$ is the affix in the masked derivative. Thus, we ignore affixes that might potentially produce equally possible derivatives such as \texttt{superwearable}.

We
use mean reciprocal rank (MRR), macro-averaged over affixes, as the
evaluation measure \citep{Radev.2002}. We calculate the MRR value of an individual affix $a$ as
\begin{equation}
\MRR_a = \frac{1}{|D_a|} \sum_{i \in D_a} \R_i^{-1},
\end{equation}
where $D_a$ is the set of derivatives containing $a$, and $\R_i$
is the predicted rank of $a$ for derivative $i$. We set $\R_i^{-1} = 0$ if $\R_i > 10$. Denoting with $\mathcal{A}$ the set of all affixes, the final MRR value is given by
\begin{equation}
\MRR = \frac{1}{|\mathcal{A}|} \sum_{a \in \mathcal{A}} \MRR_a.
\end{equation}

\subsection{Segmentation Methods}

\begin{table} [t!]\centering
\resizebox{\linewidth}{!}{%
\begin{tabular}{@{}lrrrrrrrr@{}}
\toprule
Method & B1 & B2  &  B3 &  B4 & B5 & B6 & B7  & $\mu\pm\sigma$\\
\midrule
HYP & \best{.197} & \best{.228} & \best{.252} & \best{.278} & \best{.300} & \best{.315} & \best{.337} & .272$\pm$.046\\ 
INIT & \second{.184} & \second{.201} & \second{.211} & \second{.227} & \second{.241} & \second{.253} & .264 & .226$\pm$.027\\ 
TOK & .141 & .157 & .170 & .193 & .218 & .245 & \second{.270} & .199$\pm$.044\\ 
PROJ & .159 & .166 & .159 & .175 & .175 & .184 & .179 & .171$\pm$.009\\ 
\bottomrule
\end{tabular}}
\caption{Performance (MRR) of pretrained BERT for prefix prediction with different segmentations. 
Best score per column in gray, second-best in light-gray.}  \label{tab:trick}
\end{table}

Since BERT distinguishes word-initial (\texttt{wear}) from
word-internal (\texttt{\#\#wear}) tokens, predicting
prefixes requires the word-internal form of the
base. However, only 795 bases
in BERT's vocabulary
have a word-internal form. Take as an example the word \texttt{unallowed}: both \texttt{un} and \texttt{allowed} are in the BERT vocabulary, but we need the token \texttt{\#\#allowed}, which does not exist (BERT tokenizes the word into \texttt{una}, \texttt{\#\#llo}, \texttt{\#\#wed}). To overcome this problem, we test the following four segmentation methods:

\begin{table*} [t!]\centering
\resizebox{\linewidth}{!}{%
\begin{tabular}{@{}lrrrrrrrrrrrrrrrr@{}}
\toprule
{} & \multicolumn{8}{c}{SHARED}  & \multicolumn{8}{c}{ SPLIT } \\
\cmidrule(lr){2-9}
\cmidrule(l){10-17}

Model & B1 & B2 & B3 & B4 & B5 & B6 & B7 & $\mu\pm\sigma$ & B1 & B2 & B3 & B4 & B5 & B6 & B7 & $\mu\pm\sigma$\\
\midrule
DagoBERT & \best{.373} & \best{.459} & \best{.657} & \best{.824} & \best{.895} & \best{.934} & \best{.957} & .728$\pm$.219 & \best{.375} & \best{.386} & \best{.390} & \best{.411} & \best{.412} & \best{.396} & \best{.417} & .398$\pm$.014\\ 
BERT+ & \second{.296} & \second{.380} & .497 & .623 & .762 & .838 & .902 & .614$\pm$.215 & \second{.303} & \second{.313} & \second{.325} & \second{.340} & \second{.341} & \second{.353} & \second{.354} & .333$\pm$.018\\ 
BERT & .197 & .228 & .252 & .278 & .300 & .315 & .337 & .272$\pm$.046 & .199 & .227 & .242 & .279 & .305 & .307 & .351 & .273$\pm$.049\\ 
LSTM & .152 & .331 & \second{.576} & \second{.717} & \second{.818} & \second{.862} & \second{.907} & .623$\pm$.266 & .139 & .153 & .142 & .127 & .121 & .123 & .115 & .131$\pm$.013\\  
RB & .064 & .067 & .064 & .067 & .065 & .063 & .066 & .065$\pm$.001 & .068 & .064 & .062 & .064 & .062 & .064 & .064 & .064$\pm$.002\\ 
\bottomrule
\end{tabular}}
\caption{Performance (MRR) of prefix (P) models. Best score per column in gray, second-best in light-gray.}  \label{tab:prefix}
\end{table*}

\begin{table*} [t!]\centering
\resizebox{\linewidth}{!}{%
\begin{tabular}{@{}lrrrrrrrrrrrrrrrr@{}}
\toprule
{} & \multicolumn{8}{c}{SHARED} & \multicolumn{8}{c}{ SPLIT } \\
\cmidrule(lr){2-9}
\cmidrule(l){10-17}

Model & B1 & B2 & B3 & B4 & B5 & B6 & B7 & $\mu\pm\sigma$ & B1 & B2 & B3 & B4 & B5 & B6 & B7 & $\mu\pm\sigma$\\
\midrule
DagoBERT & \best{.427} & \best{.525} & \best{.725} & \best{.868} & \best{.933} & \best{.964} & \best{.975} & .774$\pm$.205 & \best{.424} & \best{.435} & \best{.437} & \best{.425} & \best{.421} & \best{.393} & \best{.414} & .421$\pm$.014\\ 
BERT+ & \second{.384} & \second{.445} & .550 & .684 & .807 & .878 & .921 & .667$\pm$.197 & \second{.378} & \second{.387} & \second{.389} & \second{.380} & \second{.364} & \second{.364} & .342 & .372$\pm$.015\\ 
BERT & .229 & .246 & .262 & .301 & .324 & .349 & .381 & .299$\pm$.052 & .221 & .246 & .268 & .299 & .316 & .325 & \second{.347} & .289$\pm$.042\\ 
LSTM & .217 & .416 & \second{.669} & \second{.812} & \second{.881} & \second{.923} & \second{.945} & .695$\pm$.259 & .188 & .186 & .173 & .154 & .147 & .145 & .140 & .162$\pm$.019\\ 
RB & .071 & .073 & .069 & .068 & .068 & .068 & .068 & .069$\pm$.002 & .070 & .069 & .069 & .071 & .070 & .069 & .068 & .069$\pm$.001\\ 
\bottomrule
\end{tabular}}
\caption{Performance (MRR) of suffix (S) models. Best score per column in gray, second-best in light-gray.}  \label{tab:suffix}
\end{table*}

\begin{table*} [t!]\centering
\resizebox{\linewidth}{!}{%
\begin{tabular}{@{}lrrrrrrrrrrrrrrrr@{}}
\toprule
{} & \multicolumn{8}{c}{SHARED}  & \multicolumn{8}{c}{ SPLIT } \\
\cmidrule(lr){2-9}
\cmidrule(l){10-17}

Model & B1 & B2 & B3 & B4 & B5 & B6 & B7 & $\mu\pm\sigma$ & B1 & B2 & B3 & B4 & B5 & B6 & B7 & $\mu\pm\sigma$\\
\midrule
DagoBERT & \best{.143} & \best{.355} & \second{.621} & \best{.830} & \best{.914} & \best{.940} & \best{.971} & .682$\pm$.299 & \best{.137} & \best{.181} & \best{.199} & \best{.234} & \best{.217} & \best{.270} & \best{.334} & .225$\pm$.059\\ 
BERT+ & \second{.103} & .205 & .394 & .611 & .754 & .851 & .918 & .548$\pm$.296 & \second{.091} & \second{.128} & \second{.145} & \second{.182} & \second{.173} & .210 & .218 & .164$\pm$.042\\ 
BERT & .082 & .112 & .114 & .127 & .145 & .155 & .190 & .132$\pm$.032 & .076 & .114 & .130 & .177 & .172 & \second{.226} & \second{.297} & .170$\pm$.069\\ 
LSTM & .020 & \second{.338} & \best{.647} & \second{.781} & \second{.839} & \second{.882} & \second{.936} & .635$\pm$.312 & .015 & .019 & .026 & .034 & .041 & .072 & .081 & .041$\pm$.024\\ 
RB & .002 & .003 & .003 & .005 & .006 & .008 & .012 & .006$\pm$.003 & .002 & .004 & .003 & .006 & .006 & .007 & .009 & .005$\pm$.002\\ 
\bottomrule
\end{tabular}}
\caption{Performance (MRR) of prefix-suffix (PS) models. Best score per column in gray, second-best in light-gray.}  \label{tab:both}
\end{table*}

\textbf{HYP.} We insert a hyphen between the prefix and the base in its word-initial form, yielding the tokens \texttt{un}, \texttt{-}, \texttt{allowed} in our example. Since both prefix and base are guaranteed to be in the BERT vocabulary (Section \ref{sec:data}), and since there are no tokens starting with a hyphen in the BERT vocabulary, BERT always tokenizes words of the form prefix-hyphen-base into prefix, hyphen, and base, making this a natural segmentation for BERT.

\textbf{INIT.} We simply use the word-initial instead of the word-internal form, segmenting the derivative into the prefix followed by the base, i.e., \texttt{un}, \texttt{allowed} in our example. Notice that this looks like two individual words to BERT since \texttt{allowed} is a word-initial unit.

\textbf{TOK.} To overcome the problem of INIT, we segment the base into word-internal tokens, i.e., our example is segmented into \texttt{un}, \texttt{\#\#all}, \texttt{\#\#owed}. This means that we use the word-internal counterpart of the base in cases where it exists.

\textbf{PROJ.} We train a projection matrix
that maps embeddings of word-initial forms of bases to
word-internal embeddings.
More specifically, we fit a matrix $\hat{\mathbf{T}} \in \mathbb{R}^{m \times m}$ ($m$ being the embedding size) via least squares,
\begin{equation}
\hat{\mathbf{T}} = \argmin_{\mathbf{T} } || \mathbf{E} \mathbf{T}  - \mathbf{E}_{\#\#}||^2_2,
\end{equation}
where $\mathbf{E}, \mathbf{E}_{\#\#} \in \mathbb{R}^{n \times m}$ are the word-initial and word-internal token input embeddings of bases with both forms. We then map bases with no word-internal form and a word-initial input token embedding $\mathbf{e}$ such as \texttt{allow} onto the projected word-internal embedding $\mathbf{e}^\top \hat{\mathbf{T}}$.

We evaluate the four segmentation methods on the SHARED test data for P with pretrained BERT\textsubscript{BASE}, using its pretrained language modeling head for prediction and filtering for prefixes. The HYP segmentation method performs best (Table \ref{tab:trick}) and is adopted 
for BERT models on P and PS.

\begin{table} [t!]\centering 
\resizebox{0.5\linewidth}{!}{%
\begin{tabular}{@{}lrr@{}}
\toprule
Model & SHARED  & SPLIT \\
\midrule
DagoBERT & \best{.943} & \best{.615} \\
LSTM  & .824 & .511 \\
LSTM (V)  & \second{.830} & \second{.520} \\
\bottomrule
\end{tabular}}
\caption{Performance on \citet{Vylomova.2017} dataset. We report accuracies for comparability. LSTM (V): LSTM in \citet{Vylomova.2017}. Best score per column in gray, second-best in light-gray.}  \label{tab:vyl}
\end{table}

\subsection{Models}

\begin{table*} [t!]\centering
\resizebox{\linewidth}{!}{%
\begin{tabular}{@{}ll@{}}
\toprule
Type & Clusters \\
\midrule
Prefixes  & \makecell[l]{  $\{$\texttt{bi}, \texttt{demi}, \texttt{fore}, \texttt{mini}, \texttt{proto}, \texttt{pseudo}, \texttt{semi}, \texttt{sub}, \texttt{tri}$\}$, $\{$\texttt{arch}, \texttt{extra}, \texttt{hyper}, \texttt{mega}, \texttt{poly}, \texttt{super}, \texttt{ultra}$\}$,\\$\{$\texttt{anti}, \texttt{contra}, \texttt{counter}, \texttt{neo}, \texttt{pro}$\}$, $\{$\texttt{mal}, \texttt{mis}, \texttt{over}, \texttt{under}$\}$, $\{$\texttt{inter}, \texttt{intra}$\}$,\\$\{$\texttt{auto}, \texttt{de}, \texttt{di}, \texttt{in}, \texttt{re}, \texttt{sur}, \texttt{un}$\}$, $\{$\texttt{ex}, \texttt{vice}$\}$, $\{$\texttt{non}, \texttt{post}, \texttt{pre}$\}$}  \\
Suffixes & \makecell[l]{ $\{$\texttt{\#\#al}, \texttt{\#\#an}, \texttt{\#\#ial}, \texttt{\#\#ian}, \texttt{\#\#ic}, \texttt{\#\#ite}$\}$, $\{$\texttt{\#\#en}, \texttt{\#\#ful}, \texttt{\#\#ive}, \texttt{\#\#ly}, \texttt{\#\#y}$\}$, $\{$\texttt{\#\#able}, \texttt{\#\#ish}, \texttt{\#\#less}$\}$,\\$\{$\texttt{\#\#age}, \texttt{\#\#ance}, \texttt{\#\#ation}, \texttt{\#\#dom}, \texttt{\#\#ery}, \texttt{\#\#ess}, \texttt{\#\#hood}, \texttt{\#\#ism}, \texttt{\#\#ity}, \texttt{\#\#ment}, \texttt{\#\#ness}$\}$,\\$\{$\texttt{\#\#ant}, \texttt{\#\#ee}, \texttt{\#\#eer}, \texttt{\#\#er}, \texttt{\#\#ette}, \texttt{\#\#ist}, \texttt{\#\#ous}, \texttt{\#\#ster}$\}$, $\{$\texttt{\#\#ate}, \texttt{\#\#ify}, \texttt{\#\#ize}$\}$} \\
\bottomrule
\end{tabular}}
\caption{Prefix and suffix clusterings produced by Girvan-Newman after 4 graph splits on the DagoBERT confusion matrix. For reasons of space, we do not list clusters consisting of only one affix.}  \label{tab:clusters}
\end{table*}

All BERT models use BERT\textsubscript{BASE} and add a derivational classification layer (DCL) with softmax activation for prediction (Figure \ref{fig:setup}). We examine three BERT models and two baselines. See Appendix \ref{app:hyp-1} for details about implementation, hyperparameter tuning, and runtime.

\textbf{DagoBERT.} We finetune both BERT and DCL on DG, a model that we call DagoBERT (short for \textbf{D}erivationally \textbf{a}nd \textbf{g}eneratively \textbf{o}ptimized \textbf{BERT}). Notice that since BERT cannot capture statistical dependencies between masked tokens \citep{Yang.2019}, all BERT-based models predict prefixes and suffixes independently in the case of PS.

\textbf{BERT+.} We keep the model weights of pretrained BERT fixed and only train DCL on DG. This is similar in nature to a probing task.

\textbf{BERT.} We use pretrained BERT and leverage its pretrained language modeling head as DCL, filtering for affixes, e.g., we compute the softmax only over prefixes in the case of P.

\textbf{LSTM.} We adapt the approach described in \citet{Vylomova.2017}, which combines the left and right contexts $\mathbf{x}^{(l)}$ and $\mathbf{x}^{(r)}$ of the masked derivative by means of two BiLSTMs with a character-level representation of the base. To allow for a direct comparison with BERT, we do not use the character-based decoder proposed by \citet{Vylomova.2017} but instead add a dense layer for the prediction. For PS, we treat prefix-suffix bundles as units (e.g., \texttt{un\#\#able}).

In order to provide a strict comparison to \citet{Vylomova.2017}, we also evaluate our LSTM and best BERT-based model on the suffix dataset released by \citet{Vylomova.2017} against the reported performance of their encoder-decoder model.\footnote{The dataset is available at \url{https://
github.com/ivri/dmorph}. While \citet{Vylomova.2017} take morpho-orthographic changes into account, we only predict affixes, not the accompanying changes in orthography (Section \ref{sec:setup}).} Notice \citet{Vylomova.2017} show that providing the LSTM with the POS of the derivative increases performance. Here, we focus on the more general case where the POS is not known and hence do not consider this setting.

\textbf{Random Baseline (RB).} The prediction is a random ranking of all affixes.

\section{Results} \label{sec:results}

\subsection{Overall Performance}

Results are shown in Tables \ref{tab:prefix}, \ref{tab:suffix}, and \ref{tab:both}. For P and S, DagoBERT clearly performs best. Pretrained BERT is better than LSTM on SPLIT but worse on SHARED. BERT+ performs better than pretrained BERT, even on SPLIT (except for S on B7). S has higher scores than P for all models and frequency bins, which might be due to the fact that suffixes carry POS information and hence are easier to predict given the syntactic context. Regarding frequency effects, the models benefit from higher frequencies on SHARED since they can connect bases with certain groups of affixes.\footnote{The fact that this trend also holds for pretrained BERT indicates that more frequent derivatives in our dataset also appeared more often in the data used for pretraining BERT.} For PS, DagoBERT also performs best in general but is beaten by LSTM on one bin. The smaller performance gap as compared to P and S can be explained by the fact that DagoBERT as opposed to LSTM cannot learn statistical dependencies between two masked tokens (Section \ref{sec:experiments}).

The results on the dataset released by \citet{Vylomova.2017} confirm the superior performance of DagoBERT (Table \ref{tab:vyl}). DagoBERT beats the LSTM by a large margin, both on SHARED and SPLIT. We also notice that our LSTM (which predicts derivational categories) has a very similar performance to the LSTM encoder-decoder proposed by \citet{Vylomova.2017}.

\begin{figure*}[t!]
        \centering
        \includegraphics[width=\textwidth]{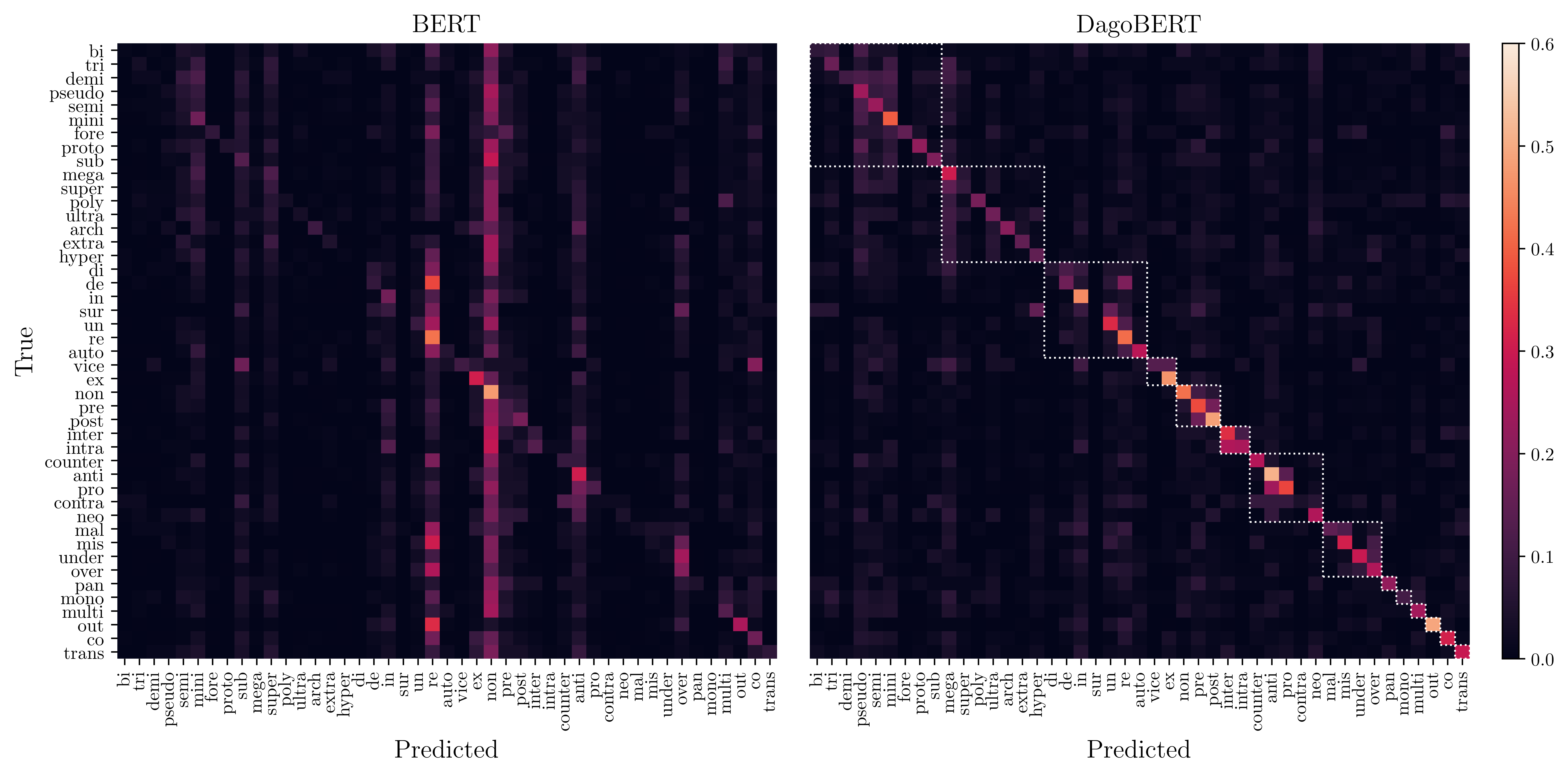}      
        \caption[]{Prefixes predicted by BERT (left) and DagoBERT (right). Vertical lines indicate that a prefix has been
 overgenerated (particularly \texttt{re} and \texttt{non} in the left panel). The white boxes in the right panel highlight the clusters produced by Girvan-Newman after 4 graph splits.}
        \label{fig:cm-pfx}
\end{figure*}

\begin{figure*}[t!]
        \centering
        \includegraphics[width=\textwidth]{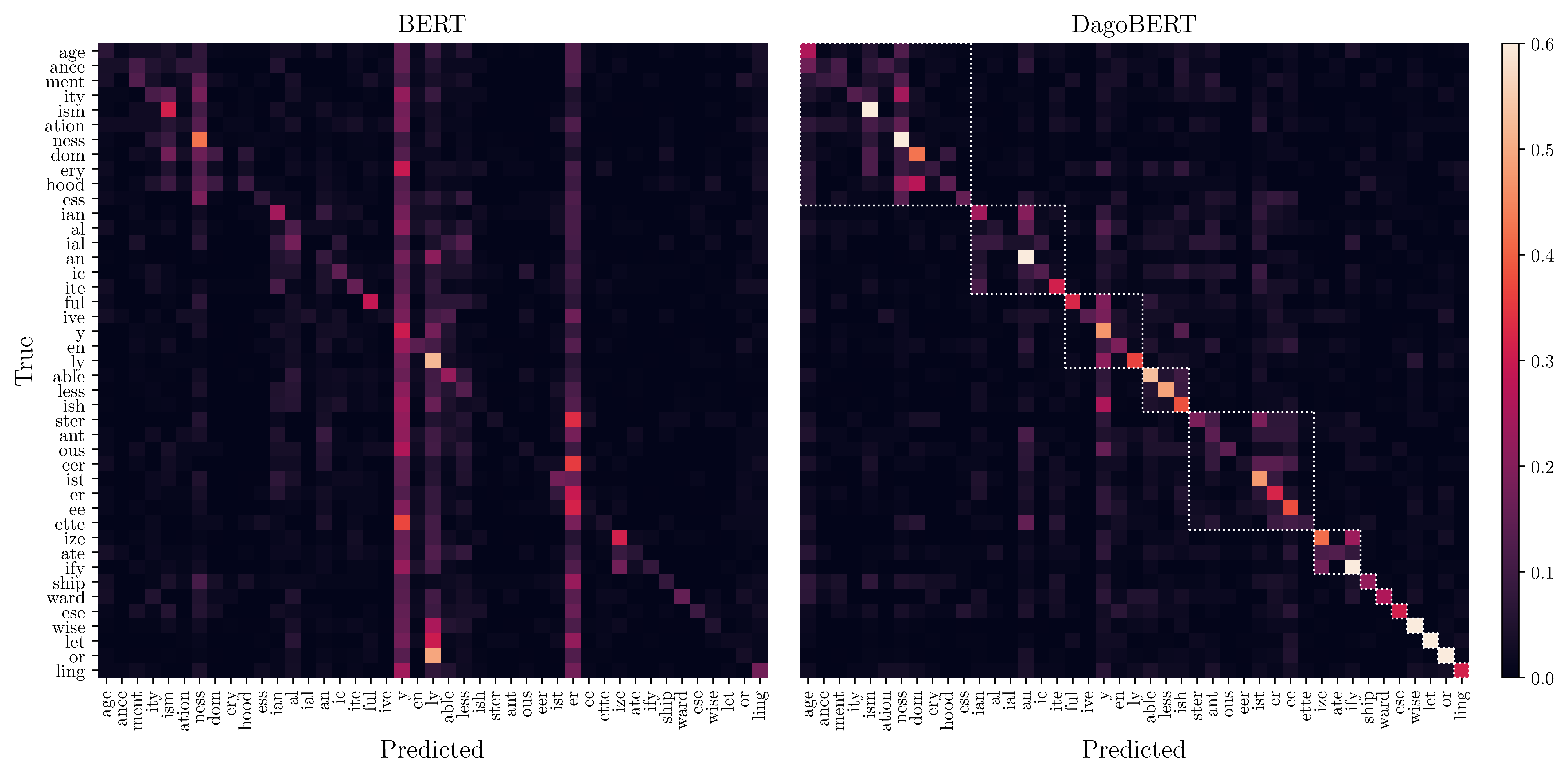}      
        \caption[]{Suffixes predicted by pretrained BERT (left) and DagoBERT (right). Vertical lines indicate that a suffix has been overgenerated (particularly \texttt{y}, \texttt{ly}, and \texttt{er} in the left panel). The white boxes in the right panel highlight the clusters produced by Girvan-Newman after 4 graph splits.}
        \label{fig:cm-sfx}
\end{figure*}

\subsection{Patterns of Confusion}

\begin{figure*}[t!]
        \centering
        \includegraphics[width=\linewidth]{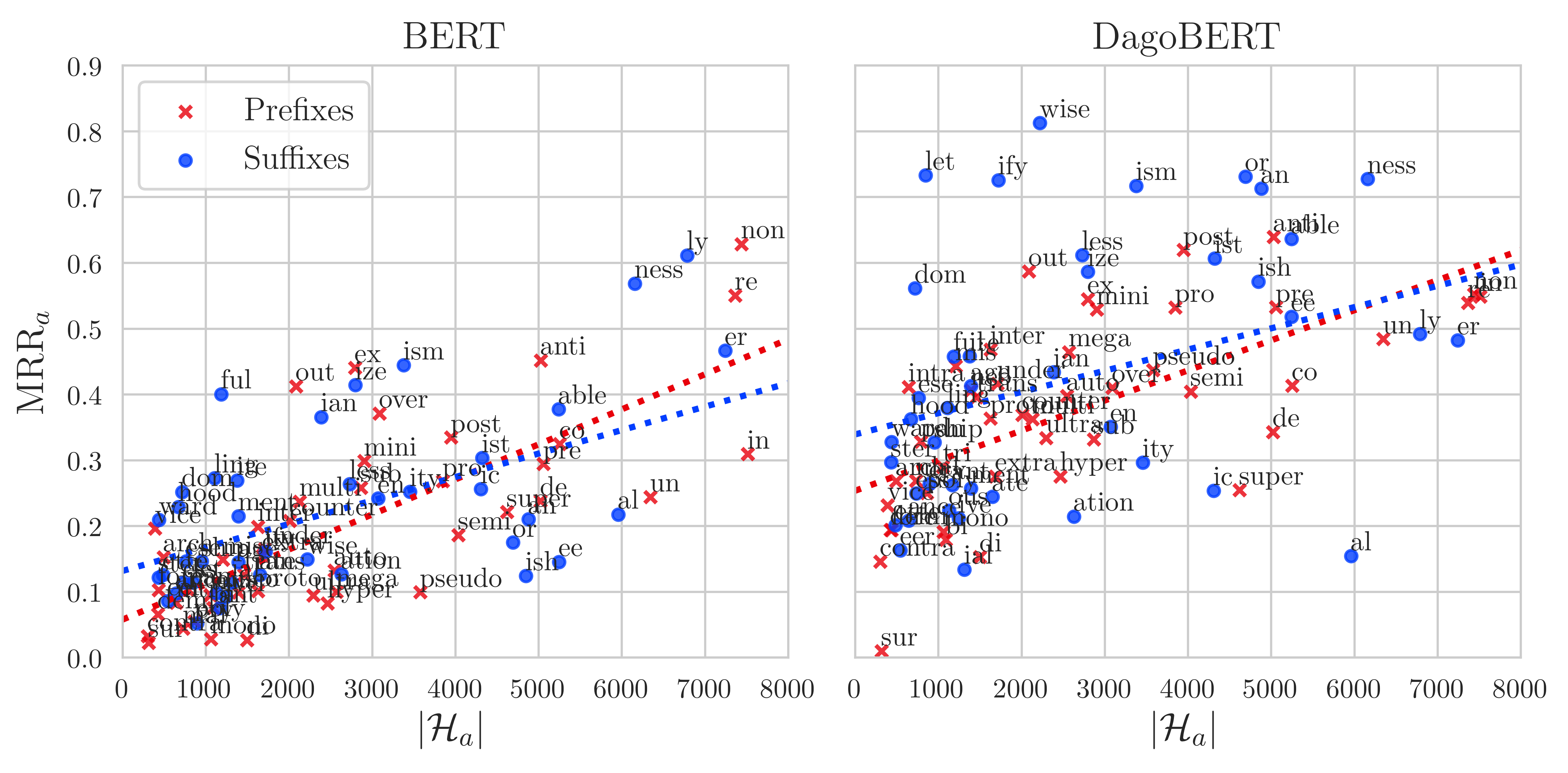}      
        \caption[]{Correlation between number of hapaxes and MRR for pretrained BERT (left) and DagoBERT (right) on B1. The highly productive suffix \texttt{y} at $(12662, 0.49)$ (left) and $(12662, 0.62)$ (right) is not shown.}
        \label{fig:correlation}
\end{figure*}

We now analyze in more detail the performance of the best performing model, DagoBERT, and contrast it with the performance of pretrained BERT. As a result of our definition of correct predictions (Section \ref{sec:setup}), the set of incorrect predictions is heterogeneous and potentially contains affixes resulting in equally possible derivatives. We are hence interested in patterns of confusion in the data. 

We start by constructing the row-normalized confusion matrix $\mathbf{C}$ for the predictions of DagoBERT on the hapax derivatives (B1, SHARED) for P and S. Based on $\mathbf{C}$, we create a confusion graph $\mathcal{G}$ with adjacency matrix $\mathbf{G}$, whose elements are
\begin{equation}
G_{ij} =  \bigl\lceil C_{ij} - \theta \bigr\rceil,
\end{equation}
i.e., there is a directed edge from affix $i$ to affix $j$ if $i$ was misclassified as $j$ with a probability greater than $\theta$. We set $\theta$ to 0.08.\footnote{We tried other values of $\theta$, but the results were similar.} To uncover the community structure of $\mathcal{G}$, we use the Girvan-Newman algorithm \citep{Girvan.2002}, which clusters the graph by iteratively removing the edge with the highest betweenness centrality.

The resulting clusters reflect linguistically interpretable groups of affixes (Table \ref{tab:clusters}). In particular, the suffixes are clustered in groups with common POS. These results are confirmed by plotting the confusion matrix with an ordering of the affixes induced by all clusterings of the Girvan-Newman algorithm (Figure \ref{fig:cm-pfx}, Figure \ref{fig:cm-sfx}). They indicate that even when DagoBERT does not predict the affix occurring in the sentence, it tends to predict an affix semantically and syntactically congruent with the ground truth (e.g., \texttt{ness} for \texttt{ity}, \texttt{ify} for \texttt{ize}, \texttt{inter} for \texttt{intra}). In such cases, it is often a more productive 
affix that is predicted in lieu of a less productive one. Furthermore, DagoBERT frequently confuses affixes denoting points on the same scale, often antonyms (e.g., \texttt{pro} and \texttt{anti}, \texttt{pre} and \texttt{post}, \texttt{under} and \texttt{over}). This can be related to recent work showing that BERT has difficulties with negated expressions \citep{Ettinger.2020, Kassner.2020}. Pretrained BERT shows similar confusion patterns overall but overgenerates several affixes much more strongly than DagoBERT, in particular \texttt{re}, \texttt{non}, \texttt{y}, \texttt{ly}, and \texttt{er}, which are among the most productive affixes in English \citep{Plag.1999b, Plag.2003}.

To probe the impact of productivity more quantitatively, we measure the cardinality of the set of hapaxes formed by means of a particular affix $a$ in the entire dataset, $|\mathcal{H}_a|$, and calculate a linear regression to predict the MRR values of affixes based on $|\mathcal{H}_a|$. $|\mathcal{H}_a|$ is a common measure of morphological productivity \citep{Baayen.1991, JanetPierrehumbert.}. This analysis shows a significant positive correlation for both prefixes ($R^2=.566$, $F(1,43) = 56.05$, $p<.001$) and suffixes ($R^2=.410$, $F(1,41) = 28.49$, $p<.001$): the more productive an affix, the higher its MRR value. This also holds for DagoBERT's predictions of prefixes ($R^2=.423$, $F(1,43) = 31.52$, $p<.001$) and suffixes ($R^2=.169$, $F(1,41) = 8.34$, $p<.01$), but the correlation is weaker, particularly in the case of suffixes (Figure \ref{fig:correlation}).

\subsection{Impact of Input Segmentation}

\begin{table*} [t!]\centering
\resizebox{\linewidth}{!}{%
\begin{tabular}{@{}lrrrrrrrrrrrrrrrr@{}}
\toprule
{} & \multicolumn{8}{c}{FROZEN}  &  \multicolumn{8}{c}{ FINETUNED } \\
\cmidrule(lr){2-9}
\cmidrule(l){10-17}
Segmentation & B1 & B2 & B3 & B4 & B5 & B6 & B7 & $\mu\pm\sigma$ &  B1 & B2 & B3 & B4 & B5 & B6 & B7 & $\mu\pm\sigma$ \\
\midrule
Morphological & \best{.634} & \best{.645} & \best{.658} & \best{.675} & \best{.683} & \best{.692} & \best{.698} & .669$\pm$.022 & \best{.762} & \best{.782} & \best{.797} & \best{.807} & \best{.800} & \best{.804} & \best{.799} & .793$\pm$.015\\ 
WordPiece & .572 & .578 & .583 & .590 & .597 & .608 & .608 & .591$\pm$.013 & .739 & .757 & .766 & .769 & .767 & .755 & .753 & .758$\pm$.010\\ 
\bottomrule
\end{tabular}}
\caption{Performance (accuracy) of BERT on morphological
  well-formedness prediction with
morphologically correct segmentation versus WordPiece
tokenization. Best score per column in gray.}  \label{tab:mwf}
\end{table*}

We have shown that BERT can generate derivatives if it is provided with the morphologically correct segmentation. At the same time, we observed that BERT's WordPiece tokenizations are often morphologically incorrect, an observation that led us to impose the correct segmentation using hyphenation (HYP). 
We now examine more directly how BERT's derivational knowledge is affected by using the original WordPiece segmentations versus the HYP segmentations.

We draw upon the same dataset as for DG (SPLIT) but perform binary instead of multi-class classification, i.e., the task is to predict whether, e.g., \texttt{unwearable} is a possible derivative in the context \texttt{this jacket is \underline{\hspace{1cm}} .} or not. As negative examples, we combine the base of each derivative (e.g., \texttt{wear}) with a randomly chosen affix different from the original affix (e.g., \texttt{\#\#ation}) and keep the sentence context unchanged, resulting in a balanced dataset. We only use prefixed derivatives for this experiment.

We train binary classifiers using BERT\textsubscript{BASE} and one of two input segmentations, the morphologically correct segmentation or BERT's WordPiece tokenization. The BERT output embeddings for all subword units belonging to the derivative in question are max-pooled and fed into a dense layer with a sigmoid activation. We examine two settings: training only the dense layer while keeping BERT's model weights frozen (FROZEN), or finetuning the entire model (FINETUNED).  See Appendix \ref{app:hyp-2} for details about implementation, hyperparameter tuning, and runtime.

Morphologically correct segmentation consistently outperforms WordPiece
tokenization,
both on FROZEN and FINETUNED (Table \ref{tab:mwf}). We interpret this in two ways. Firstly, the type of segmentation used by BERT impacts how much derivational knowledge can be learned, with positive effects of morphologically valid segmentations. Secondly, the fact that there is a performance gap even for models with frozen weights indicates that a morphologically invalid segmentation can blur the derivational knowledge that is in principle available and causes BERT to force semantically unrelated words to have similar representations. Taken together, these findings provide further evidence for the crucial importance of morphologically valid segmentation strategies in language model pretraining \citep{Bostrom.2020}.

\section{Related Work}

PLMs such as ELMo \citep{Peters.2018}, GPT-2 \citep{Radford.2019}, and BERT \citep{Devlin.2019} have been the focus of much recent work in NLP. Several studies have been devoted to the linguistic knowledge encoded by the parameters of PLMs (see \citet{Rogers.2020} for a review), particularly syntax \citep{Goldberg.2019, Hewitt.2019, Jawahar.2019, Lin.2019} 
and semantics \citep{Ethayarajh.2019, Wiedemann.2019, Ettinger.2020}. There is also a recent study examining morphosyntactic information in a PLM, specifically BERT \citep{Edmiston.2020}.

There has been relatively little recent work on derivational morphology in NLP. Both \citet{Cotterell.2017} and \citet{Deutsch.2018} propose neural architectures that represent derivational meanings as tags. More closely related to our study, \citet{Vylomova.2017} develop an encoder-decoder model that uses the context sentence for predicting deverbal nouns. \citet{Hofmann.2020b} propose a graph auto-encoder that models the morphological well-formedness of derivatives.

\section{Conclusion}

We show that a PLM, specifically BERT, can generate derivationally complex words. Our best model, DagoBERT, clearly beats an LSTM-based model, the previous state of the art in DG. DagoBERT's errors are mainly due to syntactic and semantic overlap between affixes. Furthermore, we demonstrate that the input segmentation impacts how much derivational knowledge is available to BERT. This suggests that the performance of PLMs could be
  further improved if a morphologically informed vocabulary of
  units were used.

\section*{Acknowledgements}

Valentin Hofmann was funded by the Arts and Humanities
Research Council and the German Academic Scholarship
Foundation.
This research was also supported by the European Research
Council
(Grant No.\ 740516).
We thank the reviewers for their detailed and helpful comments.

\bibliography{emnlp2020}
\bibliographystyle{acl_natbib}

\appendix

\section{Appendices}

\subsection{Data Preprocessing} \label{app:preproc}

We filter the posts for known bots and spammers \cite{Tan.2015}. We exclude posts written in a language other than English and remove strings containing numbers, references
to users, and hyperlinks. Sentences are filtered to contain between 10 and 100 words. We control that derivatives do not appear more than once in a sentence.

\subsection{Hyperparameters} \label{app:hyp-1}

We tune hyperparameters on the development data separately for each frequency bin (selection criterion: MRR). Models are trained with categorical cross-entropy as the loss function and Adam \citep{Kingma.2015} as the optimizer. Training and testing are performed on a GeForce GTX 1080 Ti GPU (11GB).

\textbf{DagoBERT.} We use a batch size of 16 and perform grid search for the learning rate $l \in \{ \num{1e-6}, \num{3e-6}, \num{1e-5}, \num{3e-5}\}$ and the number of epochs $n_e \in \{1, \dots ,8\}$ (number of hyperparameter search trials: 32). All other hyperparameters are as for BERT\textsubscript{BASE}. The number of trainable parameters is 110,104,890.

\textbf{BERT+.} We use a batch size of 16 and perform grid search for the learning rate $l \in \{ \num{1e-4}, \num{3e-4}, \num{1e-3}, \num{3e-3}\}$ and the number of epochs $n_e \in \{1, \dots ,8\}$ (number of hyperparameter search trials: 32). All other hyperparameters are as for BERT\textsubscript{BASE}. The
number of trainable parameters is 622,650.

\textbf{LSTM.} We initialize word embeddings with 300-dimensional GloVe \citep{Pennington.2014} vectors and character embeddings with 100-dimensional random vectors. The BiLSTMs consist of three layers and have a hidden size of 100. We use a batch size of 64 and perform grid search for the learning rate $l \in \{ \num{1e-4}, \num{3e-4}, \num{1e-3}, \num{3e-3}\}$ and the number of epochs $n_e \in \{1, \dots, 40\}$ (number of hyperparameter search trials: 160). The
number of trainable parameters varies with the type of the model due to different sizes of the output layer and is 2,354,345 for P, 2,354,043 for S, and 2,542,038 for PS models.\footnote{Since models are trained separately on the frequency bins, slight variations are possible if an affix does not appear in a particular bin. The reported numbers are for B1.}

Table \ref{tab:main-stats} lists statistics of the validation performance over hyperparameter search trials and provides information about the best validation performance as well as corresponding hyperparameter configurations.\footnote{Since expected validation performance \citep{Dodge.2019} may not be correct for grid search, we report mean and standard deviation of the performance instead.} We also report runtimes for the hyperparameter search. 

For the models trained on the \citet{Vylomova.2017} dataset, hyperparameter search is identical as for the main models, except that we use accuracy as the selection criterion. Runtimes for the hyperparameter search in minutes are 754 for SHARED and 756 for SPLIT in the case of DagoBERT, and 530 for SHARED and 526 for SPLIT in the case of LSTM. Best validation accuracy is .943 ($l = \num{3e-06}$, $n_e = 7$) for SHARED 
and .659 ($l = \num{1e-05}$, $n_e = 4$) for SPLIT in the case of DagoBERT, 
and .824 ($l = \num{1e-04}$, $n_e = 38$) for SHARED and .525 ($l = \num{1e-04}$, $n_e = 33$) for SPLIT in the case of LSTM.

\subsection{Hyperparameters} \label{app:hyp-2}

We use the HYP segmentation method for models with morphologically correct segmentation. We tune hyperparameters on the development data separately for each frequency bin (selection criterion: accuracy). Models are trained with binary cross-entropy as the loss function and Adam as the optimizer. Training and testing are performed on a GeForce GTX 1080 Ti GPU (11GB).

For FROZEN, we use a batch size of 16 and perform grid search for the learning rate $l \in \{ \num{1e-4}, \num{3e-4}, \num{1e-3}, \num{3e-3}\}$ and the number of epochs $n_e \in \{1, \dots ,8\}$ (number of hyperparameter search trials: 32). The number of trainable parameters is 769. For FINETUNED, we use a batch size of 16 and perform grid search for the learning rate $l \in \{ \num{1e-6}, \num{3e-6}, \num{1e-5}, \num{3e-5}\}$ and the number of epochs $n_e \in \{1, \dots ,8\}$ (number of hyperparameter search trials: 32). The number of trainable parameters is 109,483,009. All other hyperparameters are as for BERT\textsubscript{BASE}.

Table \ref{tab:mwf-stats} lists statistics of the validation performance over hyperparameter search trials and provides information about the best validation performance as well as corresponding hyperparameter configurations. We also report runtimes for the hyperparameter search.

\begin{table*} \centering
\resizebox{\linewidth}{!}{%
\begin{tabular}{@{}llrrrrrrrrrrrrrrr@{}}
\toprule
{} & {} & {} & \multicolumn{7}{c}{SHARED}  & \multicolumn{7}{c}{ SPLIT } \\
\cmidrule(lr){4-10}
\cmidrule(l){11-17}

Model & & & B1 & B2 & B3 & B4 & B5 & B6 & B7 & B1 & B2 & B3 & B4 & B5 & B6 & B7 \\
\midrule

\multirow{16}{*}{DagoBERT} 

& \multirow{5}{*}{P}

& $\mu$ & .349&.400&.506&.645&.777&.871&.930&.345&.364&.375&.383&.359&.359&.357\\
&& $\sigma$ & .020&.037&.096&.160&.154&.112&.064&.018&.018&.018&.019&.018&.017&.022\\
& &$\max$ & .372 & .454 & .657 & .835 & .896 & .934 & .957 & .368 & .385 & .399 & .412 & .397 & .405 & .392\\ 
&& $l$ &1e-5&3e-5&3e-5&3e-5&1e-5&3e-6&3e-6&3e-6&1e-5&3e-6&1e-6&3e-6&1e-6&1e-6\\

& & $n_e$ &3&8&8&8&5&8&6&5&3&3&5&1&1&1\\

\cmidrule(l){2-17}

& \multirow{5}{*}{S}

& $\mu$ &.386&.453&.553&.682&.805&.903&.953&.396&.403&.395&.395&.366&.390&.370\\
&& $\sigma$ &.031&.058&.120&.167&.164&.118&.065&.033&.024&.020&.020&.019&.029&.027\\
&& $\max$ & .419 & .535 & .735 & .872 & .933 & .965 & .976  & .429 & .430 & .420 & .425 & .403 & .441 & .432 \\
&& $l$ &3e-5&3e-5&3e-5&3e-5&1e-5&1e-5&3e-6&3e-5&1e-5&3e-6&1e-6&1e-6&1e-6&1e-6\\

&& $n_e$ &2&7&8&6&8&7&6&2&3&5&7&3&2&1\\

\cmidrule(l){2-17}

& \multirow{5}{*}{PS}

& $\mu$ &.124&.214&.362&.554&.725&.840&.926&.119&.158&.175&.194&.237&.192&.176\\
&& $\sigma$ &.018&.075&.173&.251&.238&.187&.119&.013&.013&.011&.016&.020&.021&.018\\
&& $\max$ & .146 & .337 & .620 & .830 & .915 & .945 & .970 & .135 & .177 & .192 & .219 & .269 & .235 & .209\\
&& $l$ &1e-5&3e-5&3e-5&3e-5&1e-5&3e-5&1e-5&
1e-5&3e-6&3e-6&1e-6&1e-6&1e-6&1e-6\\

&& $n_e$ &6&8&8&5&8&3&7&6&8&3&4&4&1&1\\
\cmidrule(l){2-17}
&& $\tau$  &192&230&314&440&631&897&1,098 & 195&228&313&438&631&897&791 \\

\midrule

\multirow{16}{*}{BERT+} & 
\multirow{5}{*}{P}

& $\mu$ &.282&.336&.424&.527&.655&.764&.860&.280&.298&.318&.324&.323&.324&.322\\
&& $\sigma$ &.009&.020&.046&.078&.090&.080&.051&.011&.007&.009&.013&.009&.012&.009\\
&& $\max$ & .297 & .374 & .497 & .633 & .759 & .841 & .901 & .293 & .312 & .334 & .345 & .341 & .357 & .346 \\ 
&& $l$ &1e-4&1e-3&3e-3&1e-3&3e-4&3e-4&3e-4&1e-4&1e-4&1e-4&1e-4&1e-4&1e-4&1e-4\\

&& $n_e$ &7&8&8&8&8&8&8&5&8&7&2&4&1&1\\

\cmidrule(l){2-17}

& \multirow{5}{*}{S} 

& $\mu$ &.358&.424&.491&.587&.708&.817&.886&.369&.364&.357&.350&.337&.335&.332\\
&& $\sigma$ &.010&.018&.043&.073&.086&.072&.049&.010&.010&.010&.013&.017&.017&.009\\
&& $\max$ & .372 & .452 & .557 & .691 & .806 & .884 & .925  & .383 & .377 & .375 & .372 & .366 & .377 & .357 \\
&& $l$ &1e-4&1e-3&1e-3&1e-3&1e-3&1e-3&3e-4&1e-4&1e-4
&1e-4&1e-4&1e-4&1e-4&1e-4\\

&& $n_e$ &4&7&8&8&7&7&8&8&4&5&1&1&1&1\\ 

\cmidrule(l){2-17}

& \multirow{5}{*}{PS} 
& $\mu$ &.084&.152&.257&.419&.598&.741&.849&.083&.104&.127&.137&.158&.139&.136\\
&& $\sigma$ &.008&.024&.062&.116&.119&.099&.062&.009&.014&.015&.014&.017&.011&.008\\
&& $\max$ & .099 & .206 & .371 & .610 & .756 & .847 & .913 & .099 & .131 & .154 & .170 & .206 & .173 & .164\\ 
& &$l$ &1e-4&3e-3&3e-3&3e-3&1e-3&1e-3&
1e-3&1e-4&1e-4&1e-4&1e-4&1e-4&1e-4&1e-4\\

&& $n_e$ &7&8&8&8&8&8&8&5&3&3&1&1&1&1\\
\cmidrule(l){2-17}
&& $\tau$ & 81&102&140&197&285&406&568&80&101&140&196&283&400&563\\

\midrule

\multirow{16}{*}{LSTM} & \multirow{5}{*}{P}

& $\mu$ &.103&.166&.314&.510&.661&.769&.841&.089&.113&.107&.106&.103&.103&.116\\

&& $\sigma$ &.031&.072&.163&.212&.203&.155&.107&.019&.024&.020&.017&.010&.010&.013\\

&& $\max$ & .159 & .331 & .583 & .732 & .818 & .864 & .909  & .134 & .152 & .141 & .138 & .121 & .120 & .139 \\ 
& &$l$ & 1e-3&1e-3&1e-3&1e-3&3e-4&1e-4&3e-4&3e-4&3e-4&3e-4&3e-4&1e-4&1e-4&3e-4\\

&& $n_e$ &33&40&38&35&35&40&26&38&36&37&38&40&37&29\\

\cmidrule(l){2-17}

& \multirow{5}{*}{S}

& $\mu$ &.124&.209&.385&.573&.721&.824&.881&.108&.133&.136&.132&.132&.127&.128\\
&& $\sigma$ &.037&.098&.202&.229&.206&.162&.111&.029&.034&.027&.015&.013&.012&.012\\
&& $\max$ &.214 & .422 & .674 & .812 & .882 & .925 & .945 & .192 & .187 & .179 & .157 & .159 & .157 & .153 \\ 

&& $l$&3e-4&1e-3&1e-3&1e-3&3e-4&3e-4&3e-4&
3e-4&3e-4&3e-4&3e-4&3e-4&1e-4&1e-4\\

&& $n_e$ &40&40&37&31&37&38&39&37&38&37&39&38&39&29\\

\cmidrule(l){2-17}

& \multirow{5}{*}{PS}

& $\mu$ &.011&.066&.255&.481&.655&.776&.848&.009&.015&.025&.035&.046&.032&.071\\
&& $\sigma$ &.005&.090&.256&.301&.276&.220&.177&.003&.004&.006&.006&.008&.008&.015\\
&& $\max$ & .022 & .346 & .649 & .786 & .844 & .886 & .931 &.016 & .024 & .038 & .047 & .065 & .055 & .104\\
&&$l$ &3e-3&3e-3&3e-3&3e-4&3e-4&3e-4&3e-4&
3e-4&3e-3&3e-4&3e-4&3e-3&3e-3&3e-4\\
&& $n_e$ &38&40&39&40&33&40&39&40&39&23&32&28&15&31\\
\cmidrule(l){2-17}

&& $\tau$ & 115&136&196&253&269&357&484 &  100&120&142&193&287&352&489\\

\bottomrule
\end{tabular}}
\caption{Validation performance statistics and hyperparameter search details. The table shows the mean ($\mu$), standard deviation ($\sigma$), and maximum ($\max$) of the validation performance (MRR) on all hyperparameter search trials for prefix (P), suffix (S), and prefix-suffix (PS) models. It also gives the 
learning rate ($l$) and number of epochs ($n_e$) with the best
validation performance as well as the runtime ($\tau$) in minutes averaged over P, S, and PS for one full hyperparameter search (32 trials for DagoBERT and BERT+, 160 trials for LSTM).}  \label{tab:main-stats}
\end{table*}

\begin{table*} [t!]\centering
\resizebox{\linewidth}{!}{%
\begin{tabular}{@{}lrrrrrrrrrrrrrrr@{}}
\toprule
{} & {} & \multicolumn{7}{c}{FROZEN}  & \multicolumn{7}{c}{ FINETUNED } \\
\cmidrule(lr){3-9}
\cmidrule(l){10-16}

Segmentation & {} & B1 & B2 & B3 & B4 & B5 & B6 & B7 & B1 & B2 & B3 & B4 & B5 & B6 & B7 \\
\midrule
\multirow{6}{*}{Morphological} 

& $\mu$ &.617&.639&.650&.660&.671&.684&.689&.732&.760&.764&.750&.720&.692&.657\\
& $\sigma$ &.010&.009&.009&.008&.014&.009&.009&.016&.017&.029&.052&.067&.064&.066\\
& $\max$ & .628 & .649 & .660 & .669 & .682 & .692 & .698 & .750 & .779 & .793 & .802 & .803 & .803 & .808\\
& $l$ & 3e-4&3e-4&1e-4&1e-4&1e-4&1e-4&1e-4&1e-5&
3e-6&3e-6&1e-6&1e-6&1e-6&1e-6\\

& $n_e$ &8&5&7&7&5&3&5&4&8&5&8&3&2&1\\
\cmidrule(l){2-16}
& $\tau$ & 137&240&360&516&765&1,079&1,511& 378& 578& 866&1,243&1,596&1,596&1,793 \\

\midrule

\multirow{6}{*}{WordPiece} 

& $\mu$ &.554&.561&.569&.579&.584&.592&.596&.706&.730&.734&.712&.669&.637&.604\\
& $\sigma$ &.011&.010&.011&.012&.011&.010&.011&.030&.021&.025&.052&.066&.061&.046\\
& $\max$ & .568 & .574 & .582 & .592 & .597 & .605 & .608 & .731 & .752 & .762 & .765 & .763 & .759 & .759 \\
& $l$ &1e-4&1e-4&1e-4&1e-4&1e-4&1e-4&
1e-4&1e-5&3e-6&3e-6&1e-6&1e-6&1e-6&1e-6\\

& $n_e$ &6&8&6&2&8&2&7&3&7&6&8&3&1&1\\

\cmidrule(l){2-16}

& $\tau$ & 139& 242& 362& 517& 765&1,076&1,507  & 379& 575& 869&1,240&1,597&1,598&1,775\\

\bottomrule
\end{tabular}}
\caption{Validation performance statistics and hyperparameter search details. The table shows the mean ($\mu$), standard deviation ($\sigma$), and maximum ($\max$) of the validation performance (accuracy) on all hyperparameter search trials for classifiers using morphological and WordPiece segmentations. It also gives the 
learning rate ($l$) and number of epochs ($n_e$) with the best
validation performance as well as the runtime ($\tau$) in minutes for one full hyperparameter search (32 trials for both morphological and WordPiece segmentations).}  \label{tab:mwf-stats}
\end{table*}

\end{document}